\RequirePackage{lineno}

\documentclass[onecolumn,floatfix]{revtex4}

\usepackage{color}
\usepackage{epsfig}
\usepackage{graphicx}
\usepackage{amsmath}

\begin{document}

\title{On Theory-training Neural Networks to Infer the Solution of Highly Coupled Differential Equations}
\author{M. Torabi Rad\footnote{Corresponding author Email address: m.torabirad@access-technology.de; mtorabirad@gmail.com}, A. Viardin, and M. Apel}
\affiliation{Access e.V., Intzestr. 5, 52072 Aachen, Germany}

\begin{abstract}

Deep neural networks are transforming fields ranging from computer vision to computational medicine, and we recently extended their application to the field of phase-change heat transfer by introducing theory-trained neural networks (TTNs) for a solidification problem \cite{TTN}. Here, we present general, in-depth, and empirical insights into theory-training networks for learning the solution of highly coupled differential equations. We analyze the deteriorating effects of the oscillating loss on the ability of a network to satisfy the equations at the training data points, measured by the final training loss, and on the accuracy of the inferred solution. We introduce a theory-training technique that, by leveraging regularization, eliminates those oscillations, decreases the final training loss, and improves the accuracy of the inferred solution, with no additional computational cost. Then, we present guidelines that allow a systematic search for the network that has the optimal training time and inference accuracy for a given set of equations; following these guidelines can reduce the number of tedious training iterations in that search. Finally, a comparison between theory-training and the rival, conventional method of solving differential equations using discretization attests to the advantages of theory-training not being necessarily limited to high-dimensional sets of equations. The comparison also reveals a limitation of the current theory-training framework that may limit its application in domains where extreme accuracies are necessary. 
\end{abstract}

\maketitle

\section{Introduction}

Machine learning is a branch of artificial intelligence that utilizes statistical learning methods to enable computers to learn from data, identify patterns, and make predictions using minimal task-specific programming instructions. In the machine learning domain, neural networks are computing systems that consist of several simple but highly interconnected processing elements called neurons. These neurons map one or multiple inputs to one or multiple outputs. Each neuron has a bias and connection weights, the values of which are determined in a process called training. After a network is trained, it can be used to infer the outputs on yet-unseen inputs. 

Deep neural networks are now transforming fields such as speech recognition, computer vision, and computational medicine. We recently extended their application to the field of phase-change heat transfer. In a procedure we termed theory-training, we used a theoretical (i.e., mathematical) model to train neural networks to simulate a solidification problem. In the literature \cite{Raissi_1, Raissi_2, Raissi_3}, similar networks are also referred to as ``physics-informed neural networks''. We, however,  use the term ``Theory-Trained Neural networks'' (TTNs) because we believe the latter term correctly emphasizes that the most important part of the process (i.e., training) is performed using a theoretical model, and not, for example, experimental measurements or image data sets, which can also contain information about the underlying physics of a problem. The term ``theory-training" also allows one to distinguish these networks from those trained using external data. 

\begin{figure}[t!]
  \centering
  \includegraphics[width=0.95\textwidth]{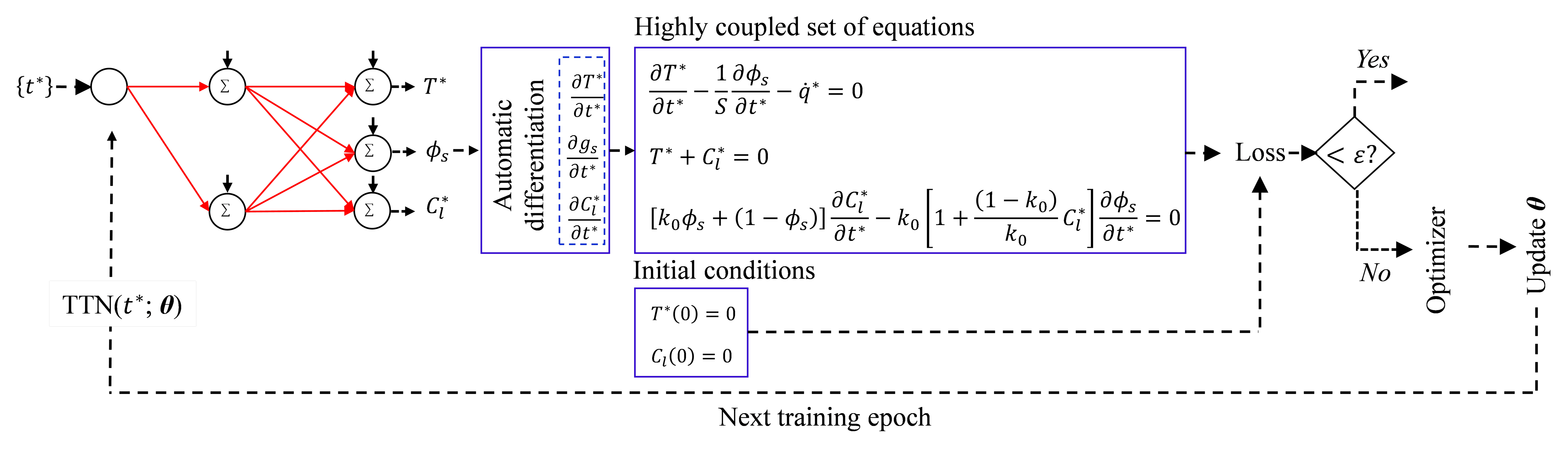}
  \caption{A schematic showing the essential elements of theory-training for a network with one
hidden layer and two nodes in that layer. The input to the network is an array containing the
times at the training points. The solid arrows (red/black) are associated with trainable network
parameters (weights/biases), and the dashed arrows show the flow of data between different steps
in the process.}
\label{fig1}
\end{figure}

The essential elements of theory-training are displayed in Figure \ref{fig1}, but discussed here only briefly; for details, the reader is referred to \cite{TTN, Raissi_1}. The figure shows a TTN with only one hidden layer and two nodes in that layer. Note that the network has eight connection weights and five biases.  The TTN is being theory-trained using a highly coupled set of equations with three unknowns and two differential equations coupled through an algebraic relation. These equations will be discussed more in Section \ref{section_eqn}. In the schematic, each solid arrow (red/black) is associated with a trainable network parameter (connection weights/node biases), and the dashed arrows show the flow of data between different elements. The input and outputs of the network are an array representing the time at the training data points and model unknowns, respectively. The derivatives of the outputs with respect to the time are calculated in a process called automatic differentiation. The outputs and their time derivatives are used to calculate the training loss, which is equal to the sum of the imbalance in the model equations and the initial conditions of the problem. If the loss is less than a small and pre-determined value, then the training ends, and the TTN can be used to infer the solution on new inputs. Otherwise, the trainable network parameters $\theta$ need to be updated, using an optimizer, such that the value of the loss is reduced.

One of the main advantages of TTNs is that they do not need any prior knowledge of the solution of the governing equations or any external data for training. They essentially self-train by relying on the ability of a neural network to learn the solution of partial differential equations (PDEs). In the literature, that ability is sometimes referred to by the term ``solving PDEs"; we, however, prefer and use the term ``inferring (or learning) the solution of PDEs" instead. The reason is, after successful training, TTNs have the power to predict the solution of a PDE at unseen data-points without actually solving the PDE, and the term ``solving PDEs" simply neglects that powerful capability. 

Theory-training is a new and active research topic and, following the pioneering works of Raissi et al. \cite{Raissi_1, Raissi_2, Raissi_3},  has so far been successfully used for learning the solution of PDEs in fields such as solid \cite{Haghighat_1, Haghighat_2} and bio \cite{bib07} mechanics, and materials science \cite{TTN}. Despite those successes, the topic is still in its infancy, and theory-training still faces numerous open questions. Some of these questions arise due to the yet-incomplete understanding of the different aspects of neural networks in the broader Deep Learning (DL) literature. For example, the dependence of training or inference performances of a neural network on the size of the network or the training data is still under debate \cite{bib08}. As another example, while some studies suggest that deep nets are preferred over the shallow ones in computer vision or speech recognition tasks, there is evidence that even single-layer fully connected feedforward nets can have a performance similar to deep nets (see \cite{bib09} and references therein). 

Another class of open questions arises because some topics being studied extensively in the broader DL literature have not gained attention in the context of theory-training yet. For example, regularization has such a central role in ML research that is rivaled in importance only by optimization \cite{GoodFellow}. Training techniques such as L2 weight regularization are commonly used to
perform computer vision or natural language processing tasks. For TTNs, however, as correctly pointed out by Raissi \cite{Raissi_4}, the effect of regularization, especially on inference accuracy is not clear. The question becomes central in comparing theory-training with a conventional and rival method of solving differential equations by discretizing the derivatives. Making any judicious statement favoring either of the methods requires one to achieve the optimal accuracy (for a given set of parameters) with both methods first and only then making the comparison. For example, comparing the predictions of a TTN that is not trained to its full learning capacity with tightly converged finite difference results is of no relevance. In a method that uses discretization, the predictions can be made as accurate as desired by simply refining the discretization steps. In theory-training, however, it is entirely unclear how the accuracy of predictions changes with the training (hyper-) parameters. 

\section{Equations}\label{section_eqn}

\subsection{A highly coupled set of equations}

The set of equations used here for theory-training represent a model for solidification of a binary alloy under an external, uniform, and constant heat extraction. The set consists of equations for conservation of energy and solute in the liquid phase and a thermodynamic constitutive relation representing the liquidus line of the phase-diagram. The only difference between this set and the one used in our previous study \cite{TTN} is in the energy equation. Here, that equation does not have any spatial derivatives and, instead, includes a constant source term, representing the heat extraction. It reads

\begin{eqnarray}\label{eqn01}
\frac{\partial T^*}{\partial t^*}=\frac{1}{S}\frac{\partial \phi_s}{\partial t^*}+\dot{q^*}
\end{eqnarray}

where $T^*$,$t^*$,$S$ and $\dot{q^*}$ represent the scaled temperature and time, Stephan number, and scaled heat
extraction rate, respectively. The equations for the scaled liquid concentration $C_l^*$ and solid fraction $\phi_s$ reader

\begin{eqnarray}\label{eqn02}
\left[k_0\phi_s+\left(1-\phi_s\right)\right]\frac{\partial C_l^*}{\partial t^*}=k_0\left[1+\frac{(1-k_0)}{k_0}C_l^*\right]\frac{\partial \phi_s}{\partial t^*}
\end{eqnarray}

and 

\begin{eqnarray}\label{eqn03}
T^*>0 &\rightarrow&\phi_s=0\\\nonumber
-1<T^*\leq 0 & \rightarrow& T^*+C^*_l=0\\
T^*\leq -1 & \rightarrow& \phi_s=1\nonumber
\end{eqnarray}

, respectively. The initial conditions reader

\begin{eqnarray}\label{eqn04}
T^*(t^*=0)=0;C_l^*(t^*=0)=0
\end{eqnarray}

Solving equations (\ref{eqn01}) to (\ref{eqn04}) using even a conventional method that discretizes the equations is not a trivial task for the following two reasons. First, one of the unknowns (i.e., $\phi_s$) does not have an explicit relation to be calculated from; therefore, a numerical scheme for updating it is required \cite{Truncated}. Second, due to the low values of the Stephan number $S$
(see the first term on the right-hand side of equation (\ref{eqn01})), there is a strong nonlinear coupling between the equations. Despite these difficulties, we have shown that theory-training can infer the solution even in the presence of spatial derivatives in the equations \cite{TTN}. Here, however, those derivatives are disregarded for the following two reasons. First, as already discussed in the introduction, the main objective of the present paper is to give in-depth insights into theory-training highly couple differential equations. That objective can be perfectly accomplished without making parts of the study not serving the objective unnecessarily more complex. Second, in the absence of spatial derivatives, we were able to derive the following analytical solution for equation (\ref{eqn01})

\begin{eqnarray}\label{eqn05}
T^*_{ex}=\frac{1}{1-k_0}\left[\frac{S\dot{q^*}(1-k_0)t^*-\sqrt{\left[S\dot{q^*}(k_0-1)t^*+Sk_0-1\right]^2+4Sk_0} - Sk_0 + 1}{2S} + k_0\right]
\end{eqnarray}

The exact analytical solutions for $C_l^*$ and $\phi_s$ were outlined in our previous work and are not
repeated here. Having an exact solution was critical in the present study because it allowed us to
quantify the $L^2$ and $H^1$ prediction errors, which are defined as \cite{bib13}

\begin{eqnarray}\label{eqn06}
||u-u_{ex}||_{L^2}=\left[\int_0^1(u-u_{ex})^2dt\right]^{0.5}
\end{eqnarray}

and 

\begin{eqnarray}\label{eqn07}
||u-u_{ex}||_{H^1}=\left[\int_0^1(u-u_{ex})^2dt+\int_0^1(\dot{u}-\dot{u}_{ex})^2dt\right]^{0.5}
\end{eqnarray}

, respectively, where $u$ can be either $T^*$,$C_l^*$ or $\phi_{S^*}$

\subsection{Standard and regularized theory-training loss}

The training loss is calculated by first calculating the errors in satisfying equations (\ref{eqn01}) to (\ref{eqn03}) at
different training data points as

\begin{eqnarray}\label{eqn08}
E_{T}^{i}=\frac{\partial T^{*,i}}{\partial t^*}-\frac{1}{S}\frac{\partial \phi_s^i}{\partial t^*}-\dot{q^*}
\end{eqnarray}

\begin{eqnarray}\label{eqn09}
E_{C_l}^i=\left[k_0\phi_s^i+\left(1-\phi_s^i\right)\right]\frac{\partial C_l^{*,i}}{\partial t^{*,i}}-k_0\left[1+\frac{(1-k_0)}{k_0}C_{l}^{*,i}\right]\frac{\partial \phi_s^i}{\partial t^{*,i}}
\end{eqnarray}

\begin{eqnarray}\label{eqn10}
E_{liq}^i=
\begin{cases}
\phi_s^i-1 & T^* \leq 0\\
\phi_s^i & T^* >0\\
T^{*,i}+C^{*,i}_l & -1<T^*\leq 0
\end{cases}
\end{eqnarray}

Similarly, the error in satisfying equation (\ref{eqn04}) is calculated from

\begin{eqnarray}\label{eqn11}
E_{ic,T}=T^*(t^*=0);E_{ic,C_l}=C_l^*(t^*=0)
\end{eqnarray}

Note that because the theoretical model here consists of ODEs, it suffices to use only one datapoint to enforce the initial conditions. Therefore, in equation (\ref{eqn11}), unlike equations (8) to (10), the error terms do not contain superscripts.
We distinguish two different losses. The first one is the standard training loss $L^S$, which is defined
as the sum of the errors associated with equations (8) to (11) and across all the training data-points.
It is calculated from 

\begin{eqnarray}\label{eqn12}
L^S=\frac{1}{N_1}\sum_{i=1}^{N_1}\left[(E_{T}^{i})^2+(E_{C_l}^{i})^2+(E^{i}_{liq})^2\right]+\left[(E_{ic,T})^2+(E_{ic,C_l})^2\right]
\end{eqnarray}

where $N_1$ is the number of training data-points that cover the interval $0<t^* \leq 1$. It is this loss
template that was minimized in our previous study \cite{TTN} and in Raissi et al. \cite{Raissi_1, Raissi_2, Raissi_3, Raissi_4}. The regularized
theory-training loss $L^R$ is defined as

\begin{eqnarray}\label{eqn13}
L^R=L^S+\gamma\sum_{i=1}^{n_h}||W^l||
\end{eqnarray}

The second term on the right-hand side represents the sum of the $L^2$ norms of the weight matrixes of the nh hidden layers. In that term, $\gamma$ and  $n_h$ are the hyper-parameters controlling the regularization strength and the number of hidden layers, respectively. Let us now discuss a subtle point about regularization (i.e., having non-zero $\gamma$ in equation (\ref{eqn13})) that becomes highly relevant in the theory-training context. In the deep learning of computer vision tasks, for example, the training loss has no physical meaning. It is only a function to fit the training data. In such tasks, following too closely patterns specific to a training dataset will degrade the network's inference performance. This is a problem known as overfitting and is commonly avoided using different regularization methods. Regularization, essentially, prevents a network from reducing the loss to the lowest values it potentially can (without regularization). Unlike computer vision tasks, in theory-training, the loss has a clear physical meaning. It represents, for example, the imbalance in the conservation of energy or matter. Imagine theory-training a single equation with a network that is large enough to make approximation errors negligible \cite{bib14} and assume that the training and ``inference" points are the same, such that there are no generalization errors. In this scenario, the $L^2$ error of the ``inferred" solution will proportional to the standard loss $L^S$.  Therefore, preventing the network from reducing $L^S$ to the lowest values it potentially can will  have a negative impact on the inference accuracy, and can be viewed as a side-effect of regularization. 
This along with other potential side-effects discussed by \cite{bib15,bib16} are, perhaps, the reasons why TTNs in the literature are usually not regularized.

To use regularization without the abovementioned side-effects, we propose to perform theorytraining in two parts:  in the first part, the network is trained by minimizing $L^R$ and, in the second part, it is trained by minimizing $L^S$. This technique can be easily implemented by calculating $\gamma$ in equation (\ref{eqn11}) from

\begin{eqnarray}\label{eqn14}
\gamma=\begin{cases}
\gamma_1 &n < n_s\\
0 &\text{otherwise}
\end{cases}
\end{eqnarray}

Because this theory-training technique regularizes the loss only during the first part of the training,
we refer to it as the Partial Regularization Theory-training Technique (PRT)

\section{Results}

All the networks theory-trained in the present study have hyperbolic tangent activation functions and were trained using the Adaptive Moment (Adam) optimizer \cite{bib17} in the first part of training (i.e., $n < n_s$) and the Sequential Least SQuares Programming (SLSQP) optimizer for the remaining epochs until convergence. Both of these optimizers are available in TensorFlow. Unless otherwise mentioned, the value of the switchover epoch $n_s$ and the number of internal training data-points were 15,000 and 200, respectively. For all the hyper-parameters, except the initial learning rate $\lambda_0$ of the Adam optimizer, the TensorFlow default values were used; the value of $\lambda_0$, as we pointed out in our previous study, is critical and is determined from the modified version of the relation we proposed in \cite{TTN} and reads as follows. The modification is to ensure that for networks with DW < 10, $\lambda_0$ will exceed 0.01. 

\begin{eqnarray}\label{eqn15}
\lambda_0=0.01\times min\left[1,\left(\frac{DW}{10}\right)^{-2}\right]
\end{eqnarray}

The results are organized as follows. In Section \ref{section_res_a}, the effects of partial regularization on the training loss and inference accuracy are investigated. The results of that section are from the base TTN. In Section \ref{section_res_b}, we discuss how the accuracy of the solution inferred by theory-training can
be improved by moving from the base TTN to the one that has the optimal inference accuracy. To better understand the reasons underlying the observations of Section \ref{section_res_b}, in Section \ref{section_res_c}, we analyze the statistics of the parameters learned by TTNs with different depths and widths. Using the findings of Section \ref{section_res_b}, we present, in Section \ref{section_res_d}, theory-training guidelines. Finally, in Section \ref{section_res_d}, we compare theory-training with a rival, conventional method that discretizes the derivatives.

\subsection{Partial Regularization Theory-training Technique (PRT)}\label{section_res_a}

\begin{figure}[t!]
  \centering
  \includegraphics[clip, trim=0.2cm 5.6cm 0.2cm 0.2cm, width=0.95\textwidth]{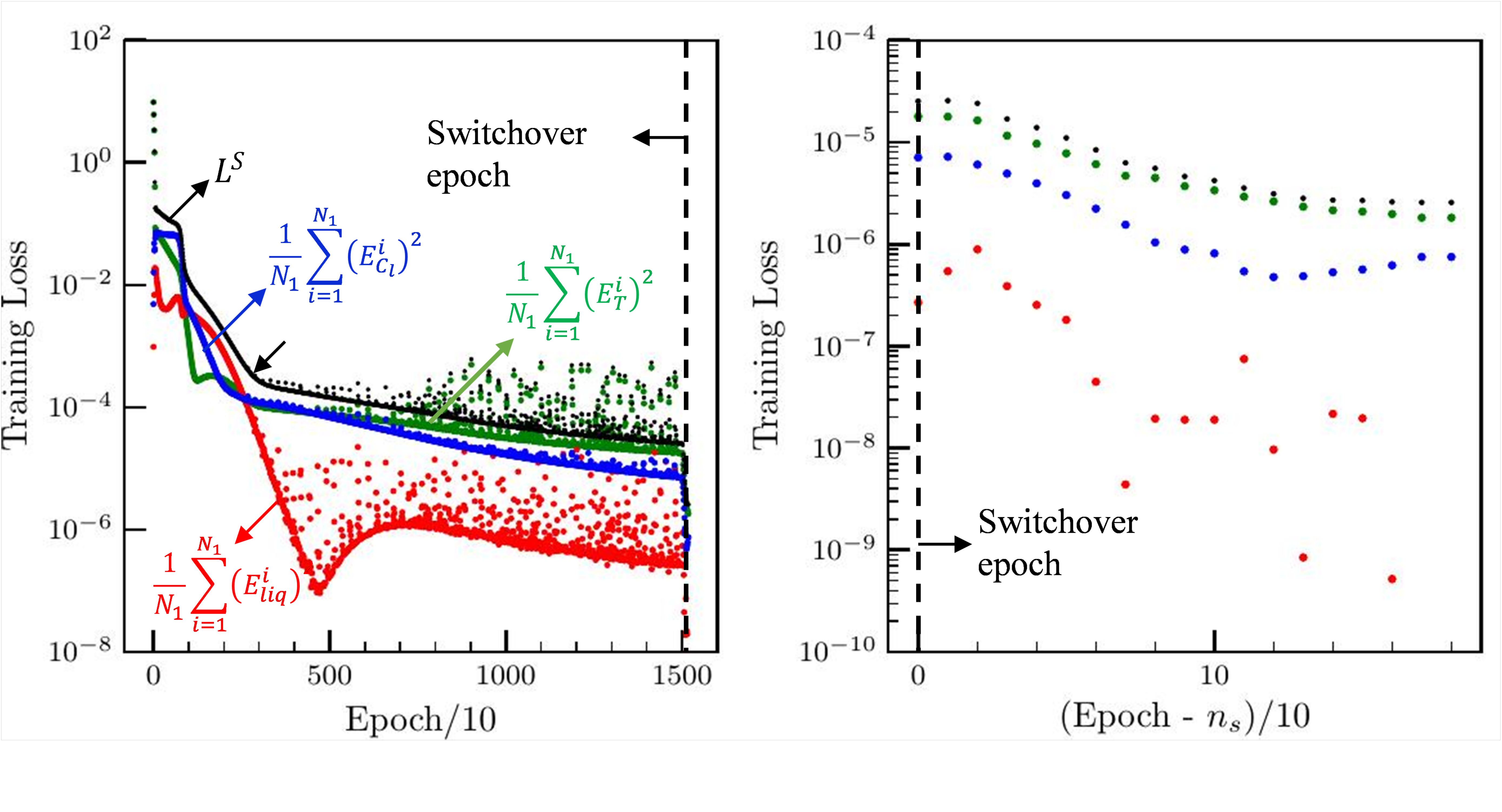}
  \caption{The total loss (black) and losses associated with different equations of the model (green, blue, and red) as a function of the training epoch, in the absence of any regularization (i.e., $\gamma_1$ in equation (\ref{eqn14})), during the first and second parts of training (left) and a close up showing the second part only (right). The dashed vertical line represents the switchover epoch. The displayed results are for the base TTN.}
\label{fig2}
\end{figure}

Figure \ref{fig2} shows the variations of the total training loss (the black curve) and the losses associated with equations (\ref{eqn08}), (\ref{eqn09}), and (\ref{eqn10}) (the green, blue, and red curves, respectively) with epochs for the base TTN and in the absence of any regularization (i.e., $\gamma_1$ in equation (\ref{eqn14})). Again, the base TTN is the one with the lowest number of hidden layers and nodes per hidden layer to reduce the loss to reasonably low values. For the current set of equations, those values were found to be $D$ =1 and $W$ = 2 (displayed schematically in Figure \ref{fig1}). The loss associated with the initial conditions (i.e., equation (\ref{eqn11})) are not displayed because they were much lower than the displayed losses. The plot on the left shows the changes during both the first and second stages of training, while the plot on the right is a close up showing only the second stage. The vertical dashed lines represent the switchover epoch, at which $\gamma$  was set from $\gamma_1$ to zero and the optimizer was switched from Adam
to SLSQP.

\begin{figure}[t!]
  \centering

  \includegraphics[clip, trim=0.2cm 4.6cm 0.2cm 0.2cm, width=0.95\textwidth]{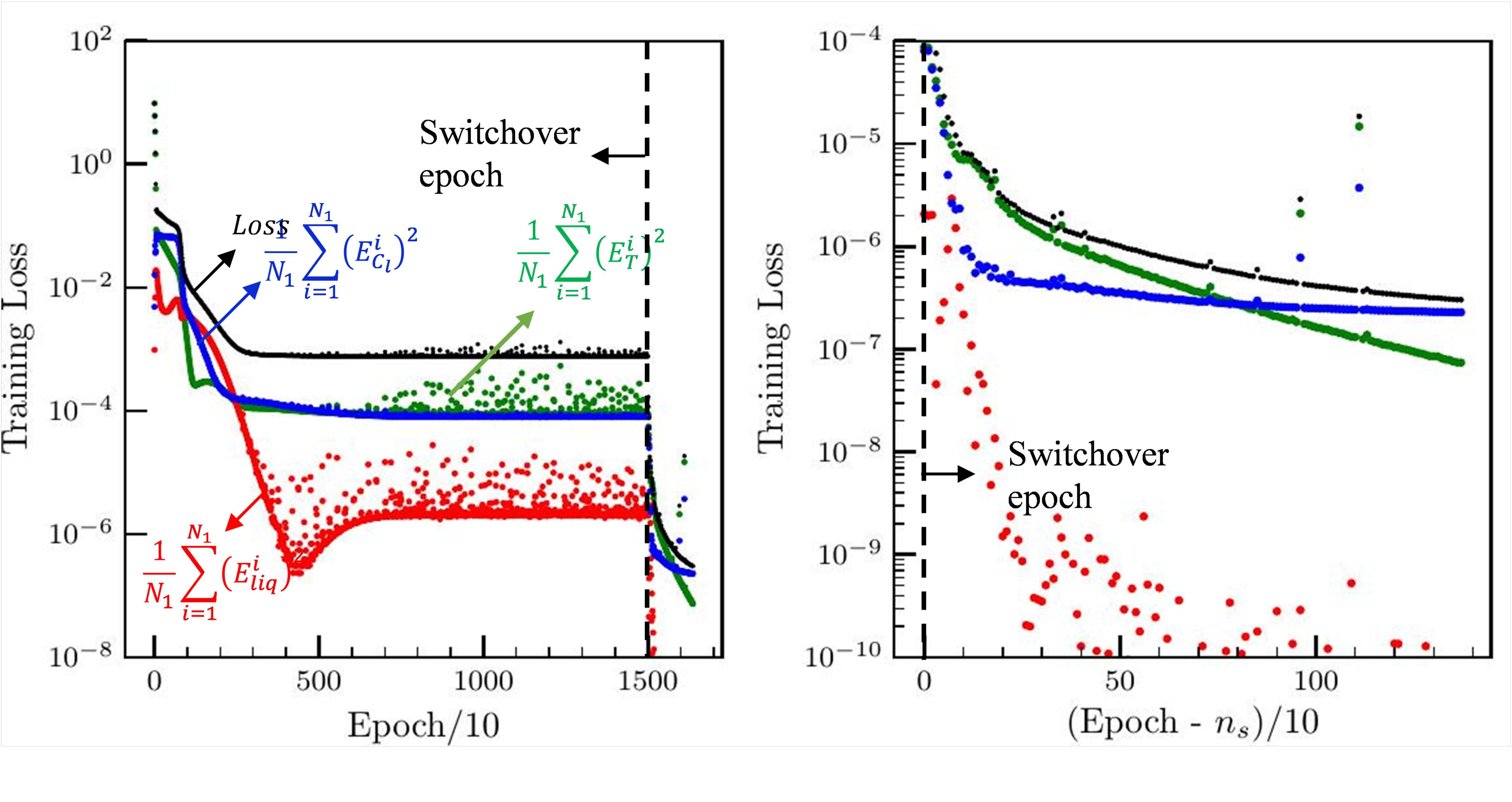}
  \caption{Results similar to those displayed in Figure \ref{fig2} but here with partial regularization (i.e.,
$\gamma_1=10^{-4}$ in equation (\ref{eqn14})). Note how partial regularization has allowed the network to reach
an oscillation-free plateau at the end of the first part of training and how the second optimizer
has become more effective in reducing the loss.}
\label{fig3}
\end{figure}

\begin{figure}[t!]
  \centering
  \includegraphics[clip, trim=0.2cm 4.6cm 0.2cm 0.2cm, width=0.95\textwidth]{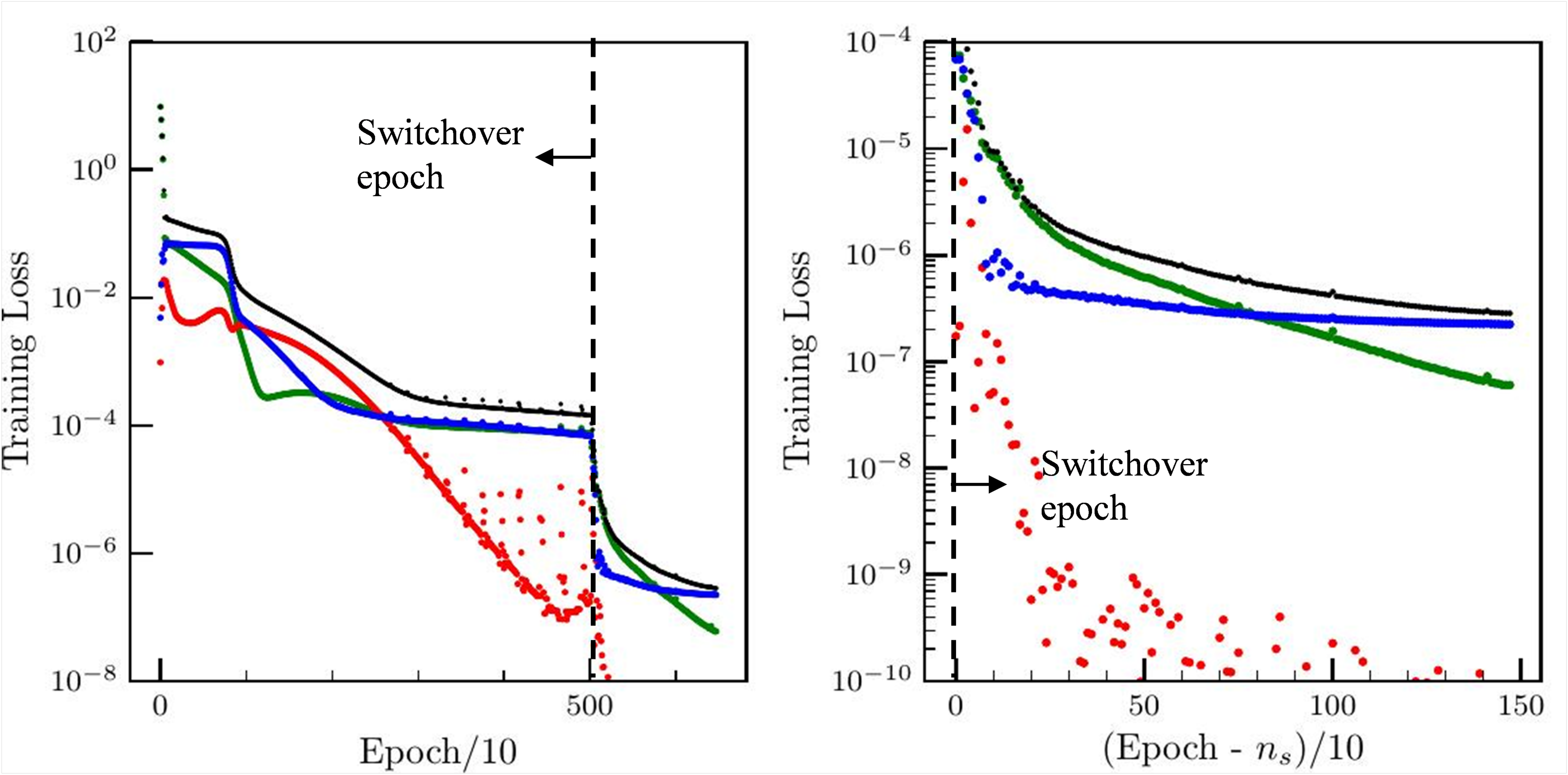}
  \caption{Results similar to those displayed in Figure \ref{fig2} but here with an earlier switchover to the second part of training (i.e., $n_s$=5000 instead of 15000)}
\label{fig4}
\end{figure}

\begin{figure}[t!]
  \centering
  \includegraphics[clip, trim=0.2cm 4.6cm 0.2cm 0.2cm, width=0.95\textwidth]{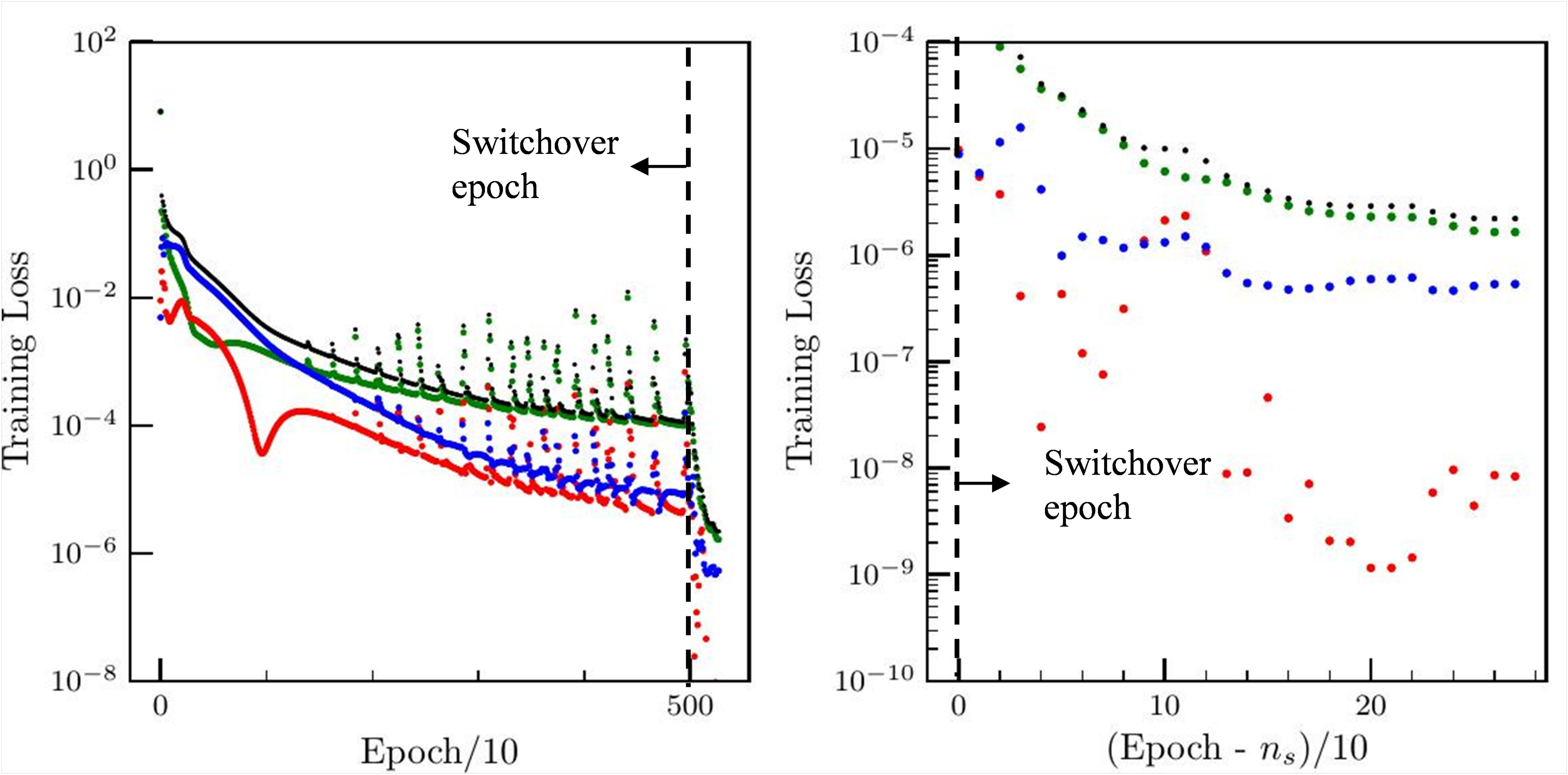}
  \caption{Results similar to those displayed in Figure \ref{fig4} but here with an earlier switchover to the second part of training (i.e., $\lambda_0$=0.05 instead of 0.01)}
\label{fig5}
\end{figure}

From the figure, it can be seen that the main contributions to the overall loss are from those
equations (within the set) that contain time derivatives (i.e., equations (\ref{eqn08}) and (\ref{eqn09})). Following the
black curve in the plot, one can note how the Adam optimizer initially decreases the total loss
monotonically and relatively quickly. At epoch around 3000 (highlighted by the downwards
inclined arrow), however, the loss starts to oscillate, and the rate of decrease in the loss suddenly
decreases. As the training continues, the oscillations become progressively stronger. These
oscillations, which can deteriorate the predictions of a TTN \cite{TTN}, are commonly observed with the
Adam optimizer \cite{bib17}. They can be eliminated using a lower initial learning rate, but that will slow
down the convergence (because it will increase the number of theory-training epochs required to
reach a converged loss). In connection with the discussion of the next figure, we will show how
PRT can eliminate them without slowing the convergence. Despite the oscillations, the average
value of loss keeps decreasing, and, at the switchover epoch (represented by the dashed vertical
line), the loss has already been reduced to $5\times10^{-5}$. After that, the second optimizer takes over (see
the right plot in the figure) and, in less than two-hundred epochs, reduces the loss further to $5\times10^{-6}$
, before the loss converges and training ends. Again, these results were in the absence of any
regularization. To investigate the effect of PRT on the training performance, we re-trained the base TTN (i.e., the
network schematically displayed in Figure \ref{fig1}) by increasing $\lambda_1$ from 0 to $10^{-4}$, while keeping
all the other hyper-parameters the same. The training curves for this new experiment are displayed
in Figure \ref{fig3}. By investigating the black curve in the figure and comparing it with the black curve in
the left plot of Figure \ref{fig2}, it can be seen that PRT causes the total loss to plateau at an epoch that is,
interestingly, almost the same as the epoch at which the oscillations in the left plot of the previous
figure started. Due to the early plateauing, the loss at the switchover epoch is about ten times higher
than it was without PRT (see Figure \ref{fig2}). In other words, the loss oscillations have been eliminated
but at the cost of having a higher loss at the switchover epoch. Nonetheless, the final loss ($\sim 2\times10^{-7}$) 
is about ten times smaller than it was without PRT ($\sim 2\times10^{-6}$ as displayed in the second plot in
Figure \ref{fig2}). This suggest that oscillations in the first part of training degrade the effectiveness of the
second optimizer in reducing the loss. That optimizer performs better when it ``receives" an
oscillation-free loss, and PRT helps to achieve that plateau. A formal analysis explaining the reason
for this observation is beyond the current understanding of how different optimizers navigate the
complex and highly non-convex loss landscape of a neural network. Nonetheless, the fact that in
the right plot of Figure \ref{fig3}, the second optimizer performs about 1,500 epochs before converging,
while in the right plot of Figure \ref{fig2}, it performs only less than a hundred epochs may be viewed as a
sign that in the latter plot, the optimizer is ``stuck" in some relatively low-quality minima.
To provide further evidence that the increase in the training performance when PRT was used is
actually due to reaching the oscillation free plateau at the end of the first part of the training, we
performed two additional training experiments both without PRT. In the first of these two
additional experiments, all the training hyper-parameters were kept the same as the ones in Figure
\ref{fig2} except the switchover epoch ns, which was reduced from 15000 to 5000. The latter epoch is
slightly after the epoch at which the oscillations in Figure \ref{fig2} started. The motivation behind 
decreasing ns was to switchover to the second optimizer before the oscillations become significant.
In the second additional experiment, we increased the learning rate (from 0.01 to 0.05) while
keeping all the other training hyper-parameters the same as the first additional experiment (so that
the oscillations become significant even at $n_s$ = 5000). The training curves for these two
experiments are displayed in Figure \ref{fig4} and Figure \ref{fig5}, respectively. By comparing the final value of
the loss, it can be seen that, again, the final value of the loss is lower (i.e., training performance is
higher) when an oscillation free plateau is reached at the end of the first part of the training.
Therefore, these two additional experiments further support the conclusion already drawn in
connection with the discussion of Figure \ref{fig2} and Figure \ref{fig3}, which stated that the second optimizer
performs better when it ``receives" an oscillation-free loss, and that plateau can be reached, without
slowing down the convergence, using PRT.

\begin{figure}[t!]
  \centering
  \includegraphics[clip, trim=0.2cm 0.2cm 0.2cm 0.2cm, width=0.95\textwidth]{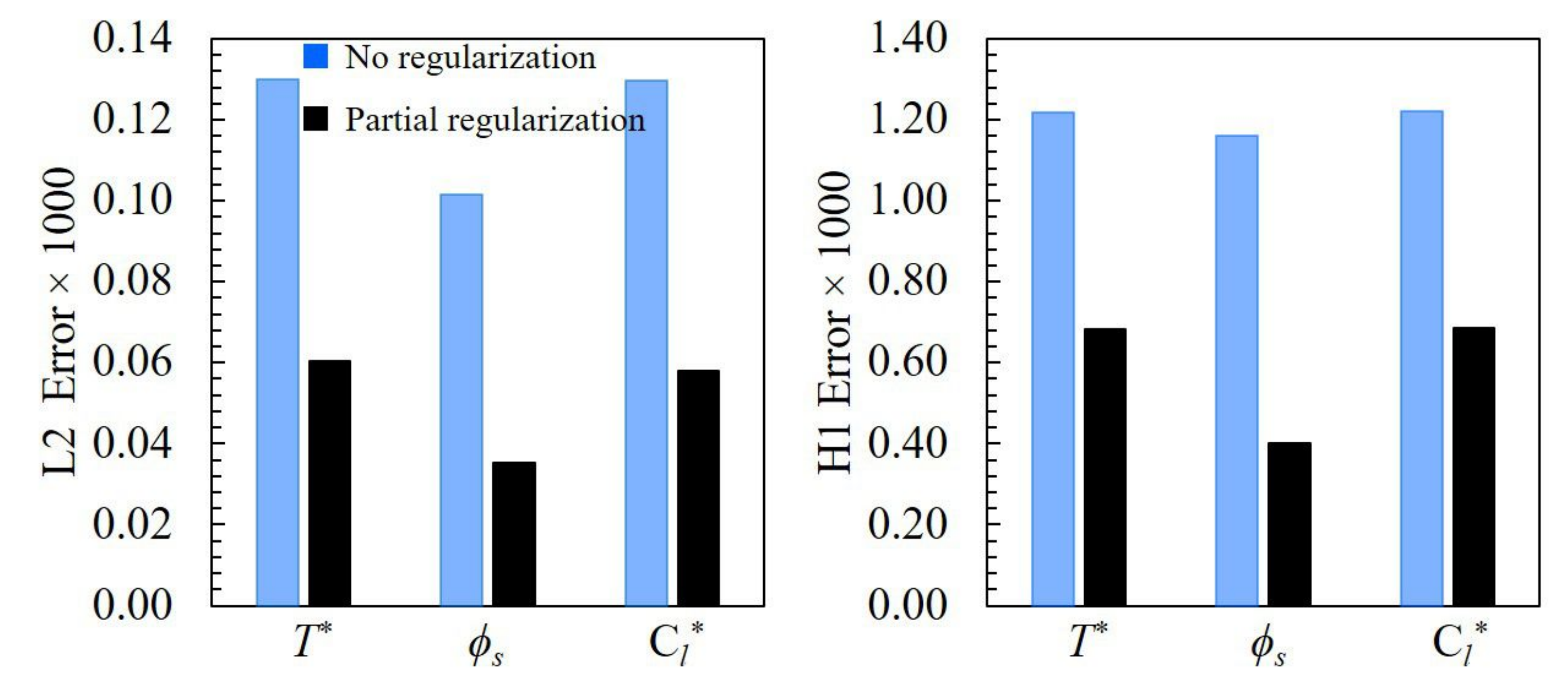}
  \caption{A comparison between the L2 (left) and H1 (right) predictions errors of the different variables (horizontal axis) predicted in the absence (blue) and presence (black) of PRT}
\label{fig6}
\end{figure}

Figure \ref{fig6} shows the $L^2$ and $H^1$ error norms (the left and rights bar plots, respectively) of the different predicted variables (see the horizontal axis) obtained from the base network when it was trained without (grey) and with (black) PRT. It is evident that PRT reduces both error norms in all the variables by about a factor of two. This improvement in accuracy is attributed to the improved training performance analyzed in connection with Figure \ref{fig3}. Note that the improvement is obtained
with no additional training cost and by simply selecting a positive $\gamma_1$ in equation \ref{eqn14}. It can also be seen that the errors in the different variables of the model have similar values. This similarity is attributed to the scaling of the equations, which has resulted in variables that are all order unity. The results presented so far were all obtained from the base TTN, which, again, is the network with the lowest number of hidden layers and nodes per hidden layer to reduce the loss to reasonably low values (i.e., satisfy the governing equations at the training points reasonably well). For the present set of equations, those values were one and two, respectively. This indicates that the solution of a model even as highly coupled as the one studied here can be learned by theory-training a network with only one hidden layer and two nodes per that layer. It is expected that for a set of equations even more complex that the ones theory-trained here, those minimum values will be higher.
Nonetheless, identifying the base TTN (i.e., finding those minimum values) should be the starting point of any theory-training practice. 

\subsection{From the base TTN to the optimal TTN to improve inference accuracy}\label{section_res_b}

\begin{figure}[t!]
  \centering
  \includegraphics[clip, trim=1.8cm 0.6cm 0.2cm 0.2cm, width=0.45\textwidth]{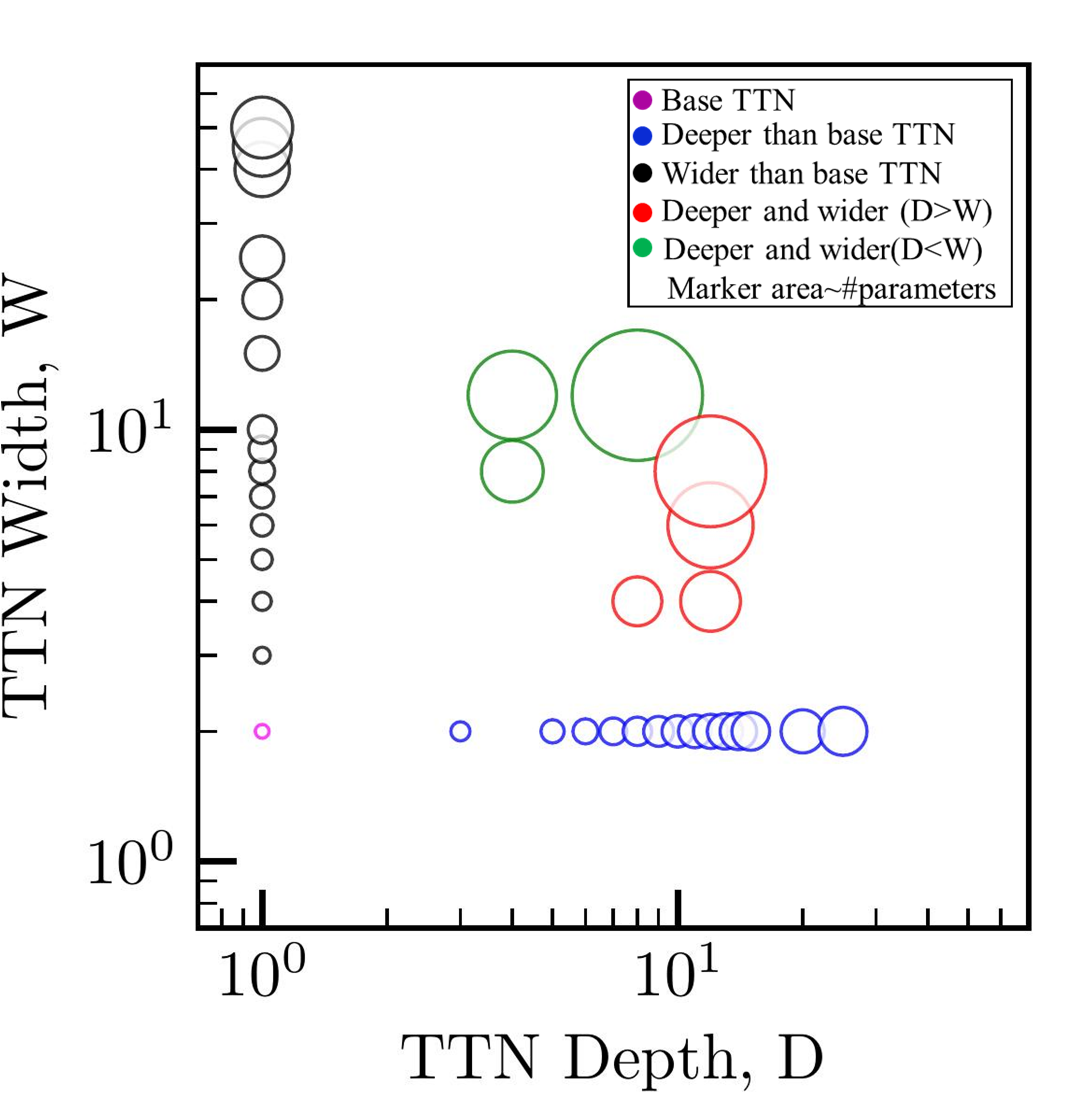}
  \caption{Depth and width of different networks theory-trained in the present study. The area of a marker is linearly proportional to the number of parameters in the corresponding TTN.}
\label{fig7}
\end{figure}

\begin{figure}[t!]
  \centering
  \includegraphics[width=0.95\textwidth]{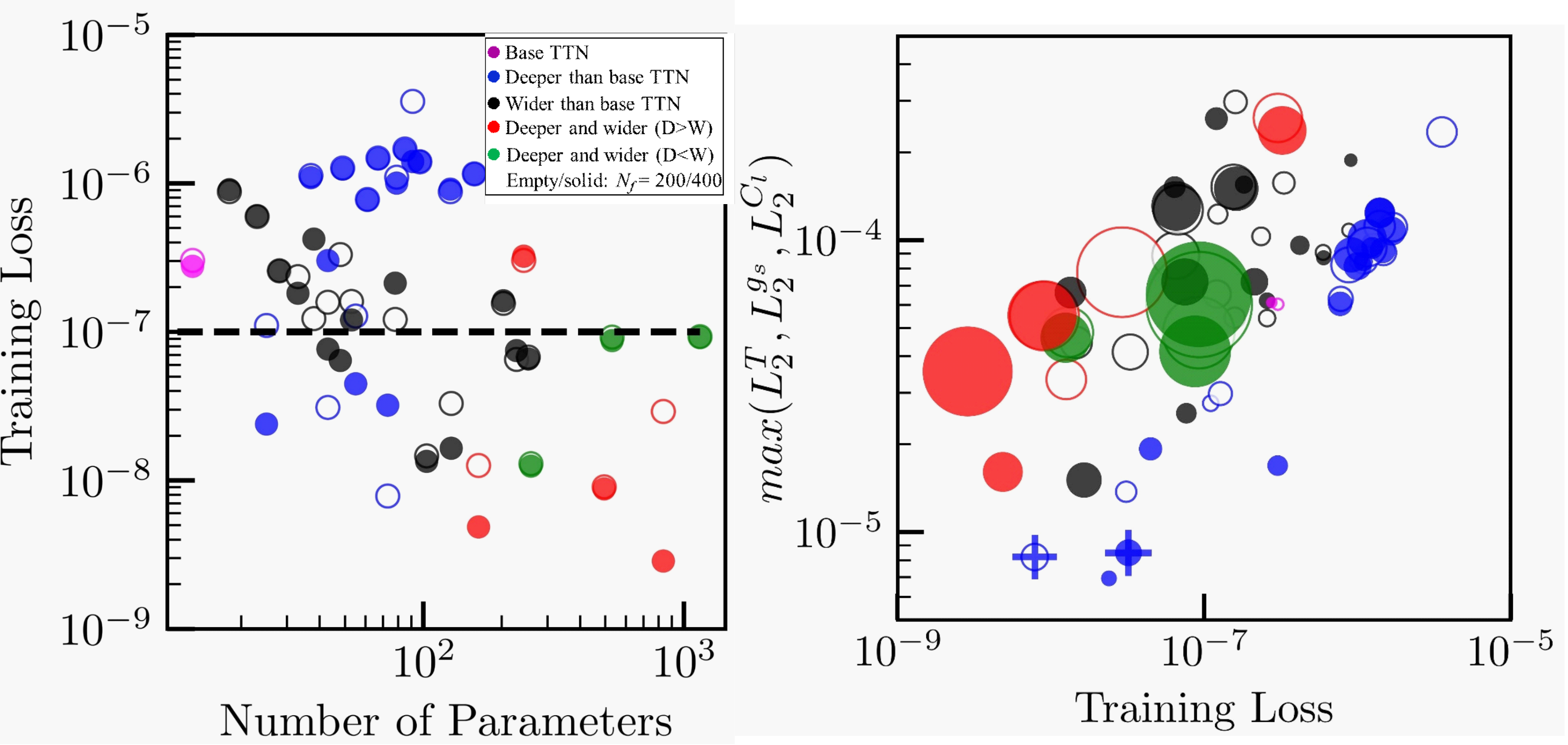}
  \caption{Training loss (left) and maximum inference error (right) of the base TTN and TTNs that are deeper and/or wider than the base (see the label). In the right plot, the size of a marker is linearly proportional to the size of the TTN the marker represents, showing that larger nets do not necessarily give more accurate predictions. Note that nets that are deeper but not wider than the base (blue markers) have the lowest error for a given loss and have a nearly linear relationship between loss and error. In the right plot, the markers distinguished by a cross represent the optimal TTN.}
\label{fig8}
\end{figure}

Once the base TTN is identified, the next natural question is whether increasing its depth and/or width will increase the inference accuracy. To address the above question, we performed theory  training experiments on more than seventy TTNs that were deeper and/or wider than the base TTN. These TTNs are represented by different markers in Figure \ref{fig7}. The base TTN is also included in the plot by the purple marker. The blue markers represent networks that are deeper but not wider than the base TTN. The black markers represent networks that are wider but not deeper than the base. Networks that are both deeper and wider than the base TTN are represented by the green (when $D < W$) and red (when $D > W$) markers, respectively. The number of trainable parameters in each network $n_p$ is equal to the sum of the number of connection weights, $W + (D-1)W^2+3W$, and the number of biases, $DW + 3$. Each TTN was trained using PRT with $\gamma_1=10^{-4}$ and with two different numbers of training points $N_f$ : $N_f$ = 200 and 4 . The switchover epoch was 15000. Experiments were performed on a PC with Intel Core i5-7500 CPU@3.40 GHz and 16.0 GB of memory. The results of these experiments are summarized in Figure \ref{fig8} and are discussed next.

Figure \ref{fig8} displays the final training loss (left) and maximum inference error (right) of TTNs that are deeper and/or wider than the base TTN (see the figure caption for the description of different colors and solid/empty markers). The left plot in the figure is analyzed first. It can be seen that for TTNs that are only wider than the base TTN (the black markers), as $n_p$ increases, the final training loss decreases initially (i.e., $n_p<30$ ) until it reaches a minimum value (here $10^{-7}$ represented by
the dashed horizontal line). After that, further increasing $n_p$ causes the loss to have only oscillations around the minimum. The initial decrease can be seen as intuitive because increasing the size of a network should increase its ability to fit the training data and should, therefore, reduce the training loss. Focusing on the blue markers in the plot shows that for TTNs that are only deeper than the base TTN, regardless of the value of $n_p$, there is no clear relationship between $n_p$ and the final training loss. Interestingly, in these networks, even a small increase in the size of the network (i.e., adding a single layer) can drastically (two orders of magnitude) change (decrease or increase) the final training loss. Also, note that the training loss of most of the TTNs that are both wider and deeper than the base TTN (red and green markers) are similar to the loss of a net that is only wider or deeper than the base. This suggests that, concerning the final training loss, networks that are deeper and wider than the base TTN have no consistent advantage over networks that are only wider or deeper than that TTN.

The right plot in Figure \ref{fig8} displays the maximum of the $L^2$ errors of the different variables ($T^*$, $\phi_s$, and $C_l^*$) as a function of the final training loss. The maximum $H^1$ errors are not displayed because their variations with the training loss were similar to the $L^2$ errors. In the plot, the size of a marker is linearly proportional to the size of the TTN the marker represents (in other words, the bigger a marker is, the larger the TTN was). It is evident that larger TTNs (i.e., larger markers) do not necessarily have lower errors, and the error of any network that is both deeper and wider than the base TTN (green/red markers) is similar to the error of a network that is only deeper or wider than that network. For example, the network with $D$ = 8 and $W$ =12 (the biggest marker in the plot, located almost at the center) has an error close to the base TTN despite having about a hundred times more parameters. This suggests that concerning the inference performance also, networks that are deeper and wider than the base TTN have no consistent advantage over networks that are only wider or deeper than that TTN. It can also be seen that networks that are deeper but not wider than the base TTN (blue markers) show two properties that make them somehow distinct from and superior to all the other nets, which are wider than the base TTN (black, green, and red markers). The former nets have the lowest prediction error for a given training loss, and their maximum error correlates almost linearly with the loss. The reason that the latter property is advantageous is discussed next. First, note that unlike the training loss, the inference errors are, in general, entirely unknown for a theory-trainer (unless equations admit an exact solution, which is rarely the case). This makes it impossible to determine
how to change $D$ or $W$ such that the error is decreased, unless the loss and errors are correlated monotonously. In the presence of such correlation, a theory-trainer can simply monitor the effect of a change in $D$ or $W$ on the training loss, and if the loss decreases due to that change, the error can also be expected to decrease. Our results suggest that networks that are deeper but not wider than the base enjoy such a correlation. Among these TTNs, the one that has the lowest loss has (almost) the lowest error, is termed the optimal TTN, and is distinguished by the cross in the right plot of the figure. Note that the optimal TTN has an inference accuracy that is about ten times better than the base TTN.

\subsection{Statistics of the learned parameters}\label{section_res_c}

The results presented in Section \ref{section_res_a} showed that the loss decreases much further after the switchover epoch. Obviously, the value of the network parameters (weights and biases) have a critical role in determining its outputs and, consequently, the loss. Therefore, the difference between the loss at the switchover and final epochs can be due to the changes in the parameters between those epochs. In addition, the results presented in Section \ref{section_res_b} showed that networks that are only deeper than the base TTN have an advantage over the ones that are wider than the base TTN. This distinction between the former and latter networks may also be linked to possible differences between the parameters that those two types of networks learn. To investigate these issues, in this section, we analyze the summary statistics and distributions of the learned
parameters.

\begin{figure}[t!]
  \centering
  \includegraphics[width=0.95\textwidth]{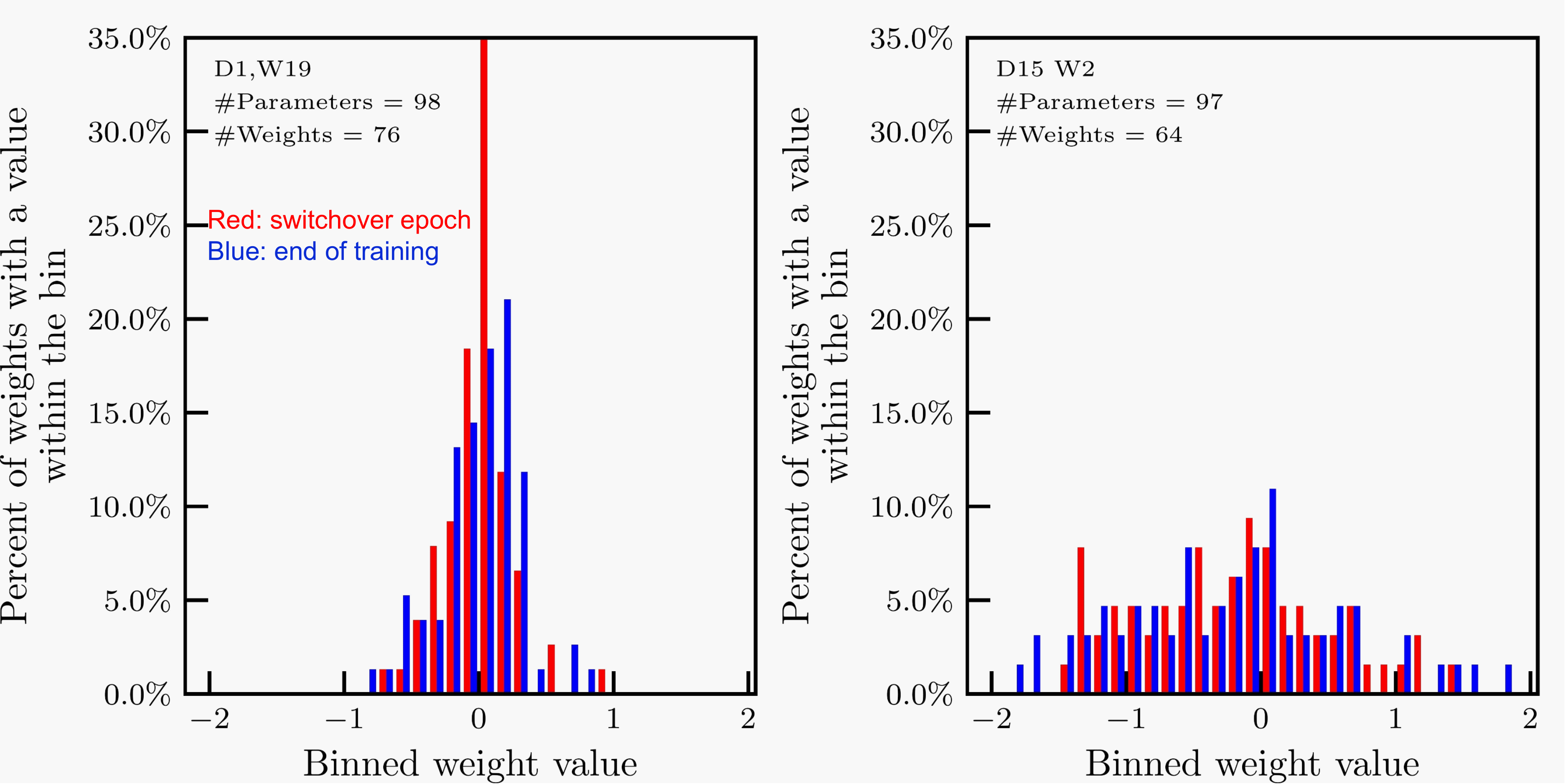}
  \caption{Comparison between the normalized distribution of the weights at the switchover (red) and final (blue) epochs for a network that is only wider (left) and only deeper (right) than the base TTN.}
\label{fig9}
\end{figure}

In figure \ref{fig9},  the normalized (with the number of weights) distributions of the weights at the switchover and final epochs are compared. The left and right panels make the comparison for a network that is wider and deeper than the base, respectively. It can be seen that, for both networks, at the final epoch, the weights have taken values that are less similar to each other than they were at the switchover epoch: the inter-variability of the weights within a network has increased after
the switchover epoch. In the left plot, that increase is the result of the post-switchover decrease in the clustering around the mean (for example, at the switchover epoch, more than thirty-five percent of the weights are essentially zero). In the right plot, however, the increase in the inter-variability is the result of post-switchover learning of a few weights with entirely different values, thereby increasing the weight range. We observed the increase in weight inter-variability for other networks
as well (nets with $D$ = 1 and $W$ = 3, $D$ = 2 and $W$ = 2, $D$ = 1 and $W$ = 7, $D$ = 5 and $W$ = 2, $D$ = 1 and $W$ = 31, $D$ = 25 and $W$ = 2), but those results are not displayed here for brevity. Therefore, we believe that the increase in the inter-variability of the weights is the reason for the order of magnitude 
further decrease in the loss after switchover observed in Section  \ref{section_res_a}. 

\begin{figure}[t!]
  \centering
  \includegraphics[width=0.95\textwidth]{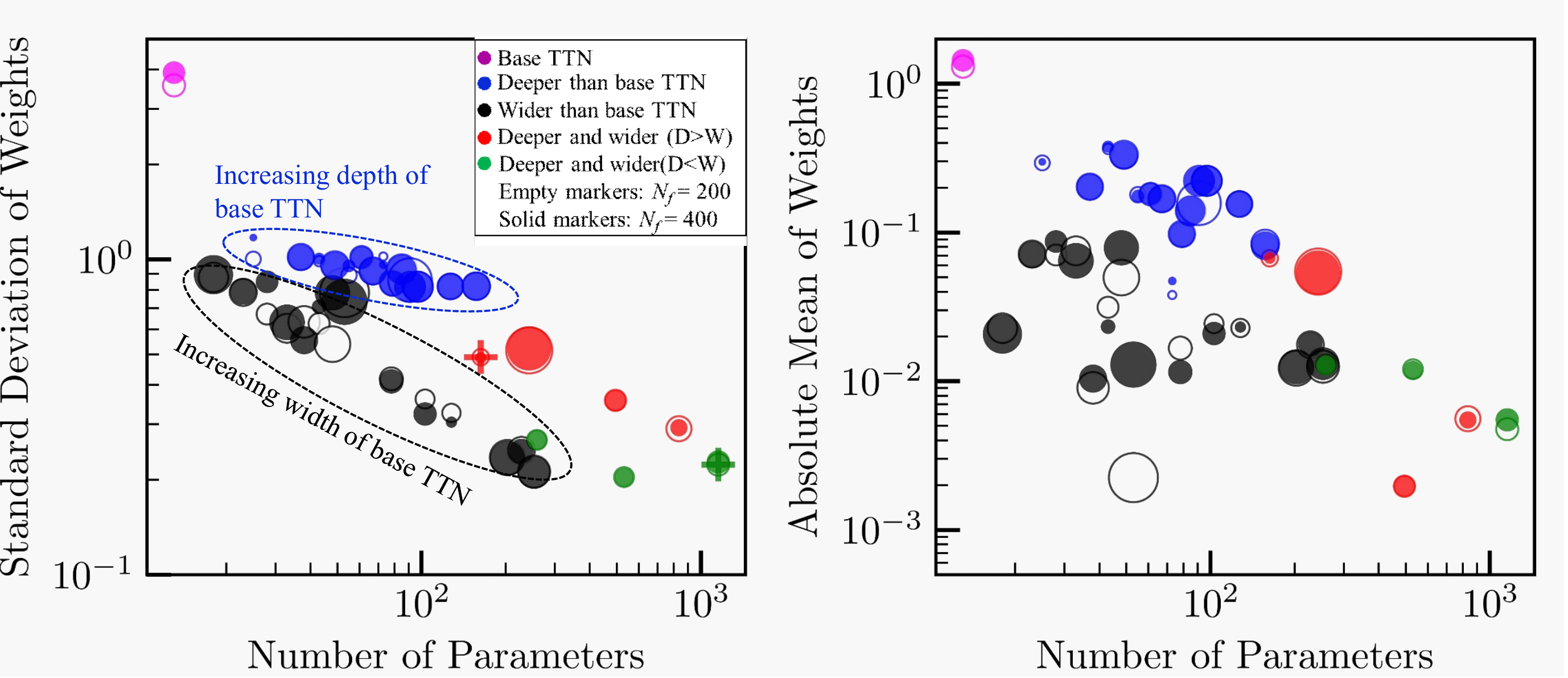}
  \caption{Standard deviation (left) and absolute mean (right) of the final weights as a function of the number of trainable parameters $n_p$ for networks resulting from going deeper and/or wider than the base TTN (see the label). Note the separation of blue markers from the black ones.}
\label{fig10}
\end{figure}

The left and right panels in Figure \ref{fig10} show, respectively, the standard deviation and mean of the final weights of the different TTNs (markers) as a function of $n_p$. In the plots, the area of a marker is selected to be linearly proportional to the maximum prediction error of the corresponding TTN (i.e., the smaller a marker is, the lower the prediction error was) to show that there is no clear relationship between the size and prediction error. It is readily evident that the blue markers are
entirely separate from the black ones. The distinction between blue and black markers was not observed in a similar plot (not included for brevity) for biases. This indicates that the only major difference between the parameters of nets that are only deeper than the base TTN with those that are wider is the higher inter-variability of the weights in the former networks. It is also interesting to note that among the networks that are both deeper and wider than the base, those with $D$ > $W$
(i.e., more deep than wide) have a higher standard deviation of weights. This suggests that among TTNs of similar size, those that are deeper are able to learn weights with relatively higher intervariability. 

\begin{figure}[tb]
  \centering
  \includegraphics[width=0.95\textwidth]{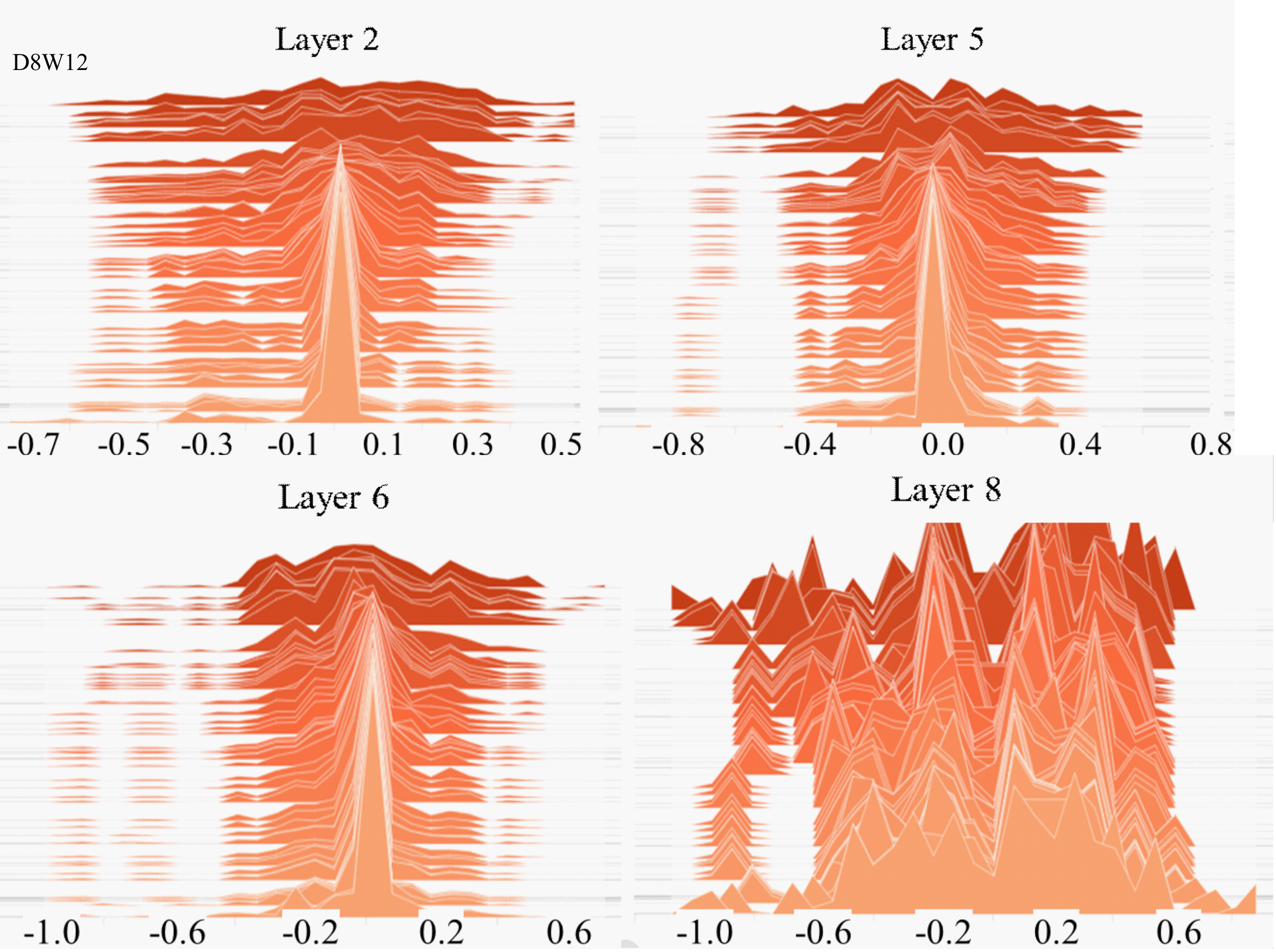}
  \caption{Epoch-evolving histograms of weights learned in different layers for the TTN with $D$= 8 and $W$ = 12.}
\label{fig11}
\end{figure}

\begin{figure}[t!]
  \centering
  \includegraphics[width=0.95\textwidth]{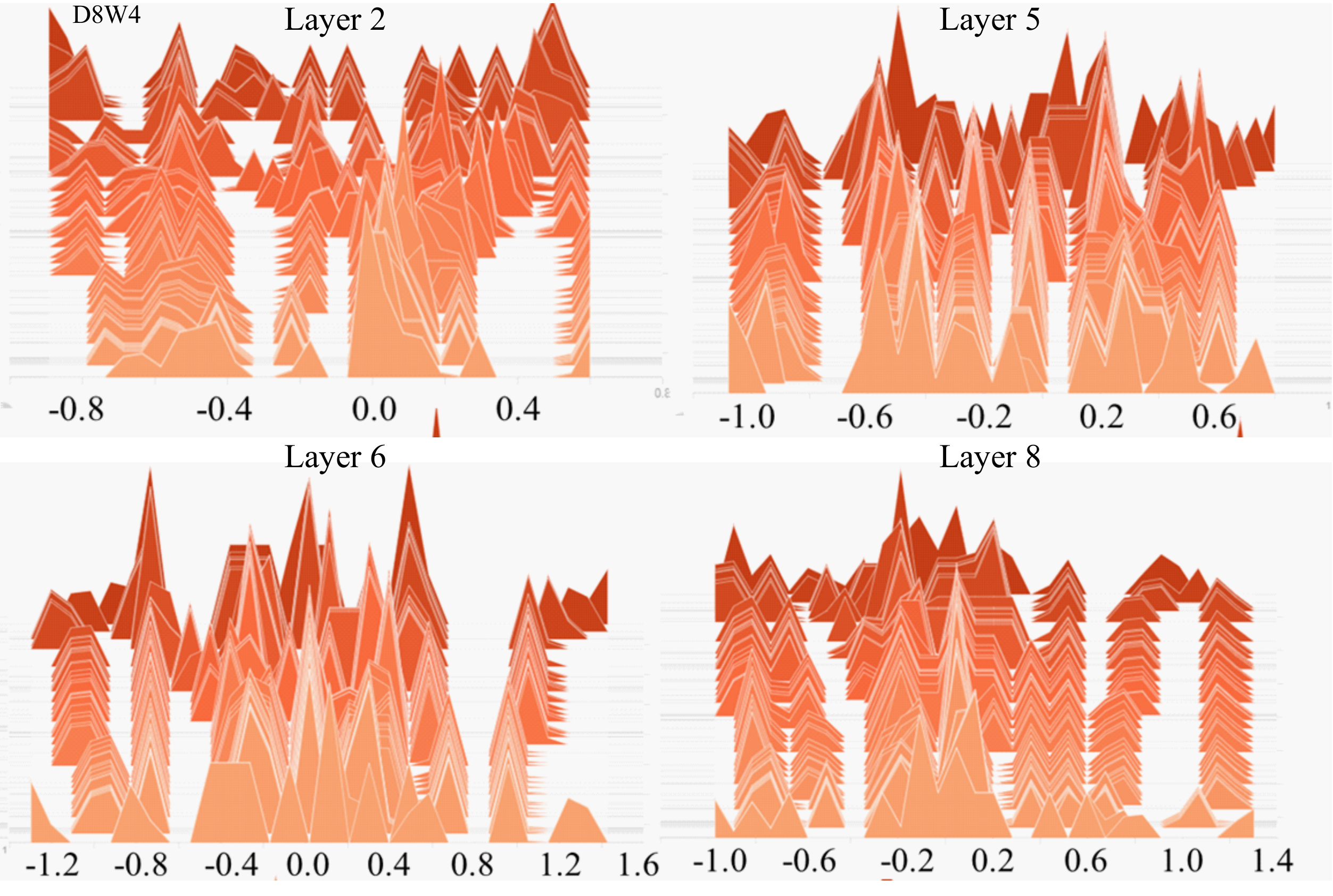}
  \caption{Epoch-evolving histograms of weights learned in different layers for the TTN with $D$= 8 and $W$ = 4.}
\label{fig12}
\end{figure}

Figure \ref{fig11} and Figure \ref{fig12} display, respectively, the variations of the weight distributions with training epochs for the TTN with $D$ = 8 and $W$ = 12 and the TTN with $D$ = 8 and $W$ = 4. These TTNs are selected for display as representatives of TTNs with $W > D$ (i.e., black and green markers in Figure \ref{fig7}) and ones with $W < D$ (i.e., blue and red markers in Figure \ref{fig7}). In Figure \ref{fig10}, these two TTNs are distinguished by a cross in the left plot. Each graph within these figures consists of slices that are stacked in the direction that appears perpendicular to the plane of view. Each slice represents a histogram where the horizontal axis shows the binned values, and the height is proportional to the percentage of the weights with a value within the bin. Slices that belong to earlier epochs are further "back" and darker, while the ones that belong to later epochs are close to the foreground and lighter in color. Figure \ref{fig11} shows that the weights of the different hidden layers evolve such that the final weights become more or less the same and all clustered around the mean zero. Such behavior is not observed in Figure \ref{fig12}. This suggests that in a network that is wide enough, adding more hidden layers simply forces the network to learn the same weight values again and again, rather than having it learn new weights.

\subsection{Theory-training guidelines}\label{section_res_d}

Based on the empirical observations analyzed in Section \ref{section_res_b}, we present the following theorytraining guidelines. Starting from a small network, find the base TTN by gradually increasing the depth and with such that the final training loss becomes reasonably low. For the set of equations we theory-trained here, the base TTN had only one hidden layer and two nodes per that layer. If the complexity of the equations is increased even further, that TTN can be expected to have a larger size; nonetheless, it can still be easily identified because any TTN smaller than the base will have a much higher training loss. Train the base TTN (and all the others that will follow) in two parts, where the loss is regularized only in the first part (see Section \ref{section_res_a} for details). Then, train networks that have more hidden layers than the base TTN. In increasing the number of hidden layers, note that adding a layer can drastically change (both decrease and increase) the training loss. Among the TTNs that are deeper but not wider than the base, the one that has the lowest loss (i.e., the optimal TTN) can be expected to have the most accurate predictions and should be used for inference. The reason for prioritizing adding layers to adding more neurons per existing layers (i.e., going deeper over going wider) is not that the former is more likely to reduce the errors. It is only because networks that are only deeper than the base have a monotonous correlation between the loss and the inference error. Therefore, for these networks, decreasing the training loss, which is a quantity that is always available to the practitioner, can be expected to decrease the inference errors; those errors are generally not available (unless the equations admit an exact solution, which is rarely the case). These guidelines are intended to ensure that the TTN that is ultimately used for inference is trained to its full realizable learning capacity, provides the most accurate predictions among TTNs that have different sizes, and does not have additional, unnecessary neurons. 

Before proceeding, we need to mention that the above guidelines are based only on the theorytraining experiments we performed here. Therefore, a necessary future study is to test their validity under different training conditions, for example, where a different set of equations is theory-trained. Note that such a set will still need to admit an exact solution so that the inference errors can be quantified. If the new set of equations is more complex, finding the base TTN may require more trial and error (because one will have to increase both D and W). However, after that TTN is identified, we expect the guidelines to be reliable, mainly because all the individual equations are lumped into a single scalar loss, and that loss is the only interface between the network and equations. In addition, in the next section, we will show that the 10X increase in the inference accuracy by switching from the base to the optimal TTN (see the last paragraph in Section \ref{section_res_b}) does not depend on $N_f$. This provides some confidence that, under training conditions different than the ones explored here, the guidelines will still be reliable if not entirely accurate. For example, there may be scenarios where among the TTNs that are deeper but not wider than the base, the one that has the lowest training loss does not have exactly the best inference accuracy. Clarifying these issues obviously requires further research. Nonetheless, the value of the present guidelines and the critical role they can play in making any judicious statement concerning the comparison between theory-training and rival methods of solving differential equations is further highlighted in the next section.

\subsection{Theory-training guidelines}\label{section_res_e}

It is known that theory-training a high-dimensional set of equations will be significantly faster than solving those equations using conventional methods for solving differential equations that discretize the equations, because the latter suffer from the so-called curse of dimensionality \cite{Raissi_5}. For a low-dimensional set of equations, however, it is entirely unclear whether theory-training has any advantages over those methods. The equations in the present study are a good candidate to
address that issue (because they are one-dimensional in the spatio-temporal domain). To that end, we solved equations (\ref{eqn01}) to (\ref{eqn03}) by discretizing the time derivatives using the backward Euler method and using the solid fraction updating scheme of Torabi Rad and Beckermann \cite{Truncated}. Again, such a
scheme is necessary because $\phi_s$ in the equations does not have an explicit relation to be calculated from. Due to the high coupling between the equations, within each time step, they had to be solved iteratively until all the variables converged (i.e., did not change with further iterations within a user-prescribed criterion). These simulations were performed using a simple (70 lines of code) Python script.

\begin{figure}[t!]
  \centering
  \includegraphics[clip, trim=1cm 1cm 1cm 1cm, width=0.95\textwidth]{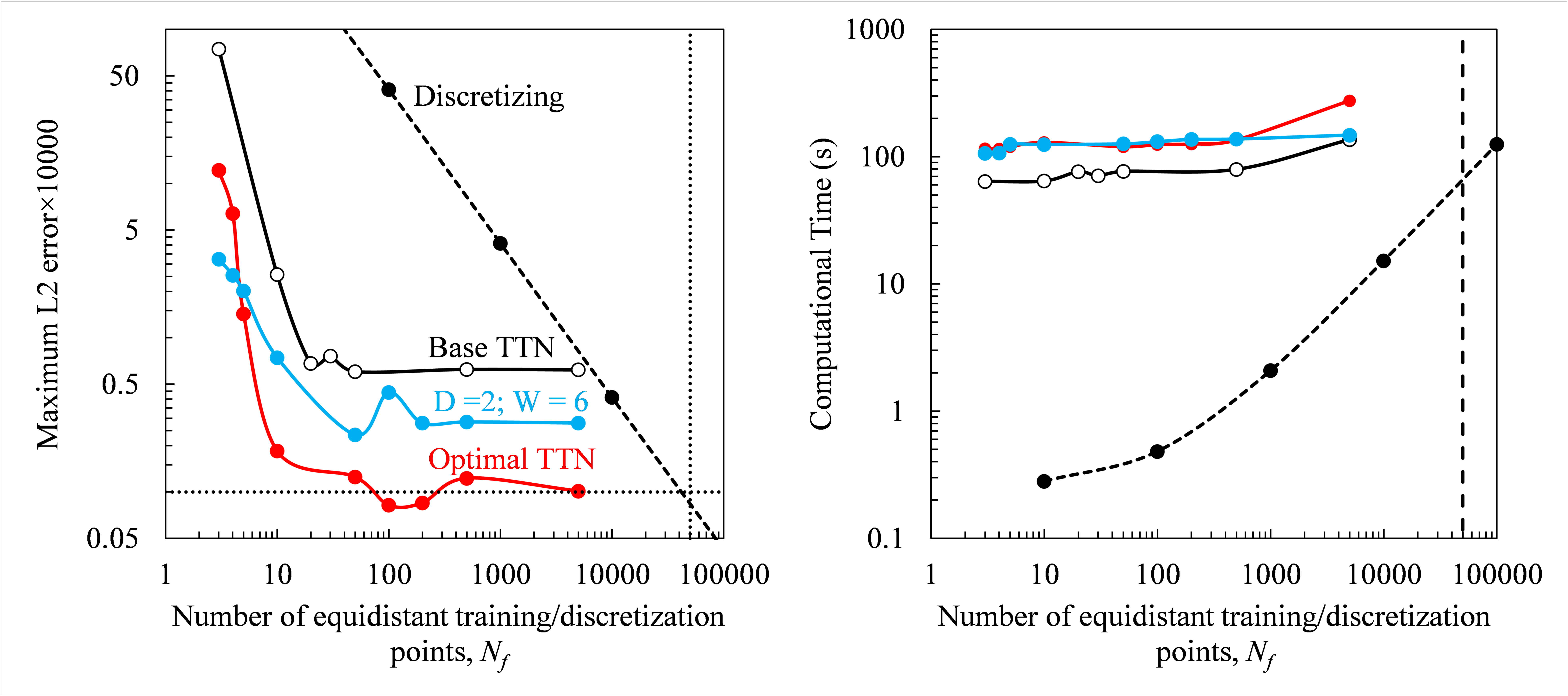}
  \caption{Comparison between the maximum $L^2$ errors (left) and computational times (right) of theory-training (solid curves) and a rival method for solving differential equations that discretizes the derivatives (dashed curves), showing that even on a one-dimensional set of equations, theory-training can be as fast as a discretizing method.}
\label{fig13}
\end{figure}

Figure \ref{fig13} displays the maximum L2 errors (left plot) and computational times (right plot) of theory-training (solid curves) and the method using discretization (dashed curves) as a function of the number of training/discretization points $N_f$ . The latter method will be referred to as the conventional method. The computational times of theory-training were almost equal to the training times because the inference was nearly instantaneous. Theory-training results are displayed for the base TTN
(black curves) and the optimal TTN (red curves), and the TTN with $D$ = 2, and $W$ = 6 (blue curves). The latter TTN has the same number of parameters as the optimal TTN. Before comparing the results of theory-training and conventional methods, it should be mentioned that the 10X improvement in the inference accuracy that was obtained by switching from the base to the optimal TTN (see the right plot in Figure \ref{fig08} and the discussion at the end of Section \ref{section_res_b}) persists regardless of the value of $N_f$ . This provides some confidence that the guidelines introduced in Section \ref{section_res_d}, which were based on observations with N f = 200 and 400 only, do not depend on N f and are general. Also, note that the training time of the optimal TTN is higher than the base TTN by less than thirty percent, indicating that the 10X improvement in the inference accuracy is obtained with no significant increase in the computational cost. It would be interesting to see whether such a gain in the inference accuracy exists for a set of equations that contain spatial derivatives, especially in the presence of sharp gradients. For such a system, if the gain turns out to be significant, it \emph{may} reflect itself by somehow alleviating a reported shortcoming in theory-training that concerns having overly-diffuse interfaces, instead of ones that should be as sharp as possible \cite{Haghighat_2}, and this is another issue that requires further research.

Comparing the red and dashed-black curves in the figure shows that the computational time required with theory-training to reach the lowest error achievable by the optimal TTN (represented by the horizontal dotted line in the left plot) is similar to the corresponding time with the conventional method: one hundred seconds compared to eighty seconds. The minor difference between the two times should not distract because these computations were performed using different computer codes (TensorFlow vs. a simple Python script). The important point to consider here is that the times are of the same order of magnitude even through the equations are only one-dimensional (in the spatio-temporal domain). Therefore, if the dimensionality of equations is increased by only one, then theory-training can be expected to clearly defeat the conventional method. This is because such an increase will require increasing $N_f$ (number of training/discretization points). That increase will increase (almost linearly) only the computational time of the conventional method (see the right plot and note that a thousand-fold increase in $N_f$ has almost no impact on the theory-training times). The similarity (between the computational times) that results in somehow unexpected result that even on a low-dimensional set of equations theory-training can have advantages over rival, conventional methods is attributed to the high coupling between the equations studied here. In the presence of such coupling, and when the equations are discretized, at each time step, numerous iterations have to be performed for the variables to converge. Before proceeding, note that if the base TTN were used in the comparison with the conventional method, the result would have been an erroneous and misleading conclusion that, on a low-dimensional set of equations, theory-training cannot have any computational advantage over discretizing, as the latter can reach the minimum error achievable by the base TTN more than ten times faster. This again highlights the importance of reaching the optimal computational time and inference accuracy, using, for example, the guidelines in Section \ref{section_res_d} before making any comparison against the rival methods.

From the left plot, it can also be seen that, while discretization errors decrease linearly as $N_f$ increases, the theory-training errors decrease with increasing $N_f$ only initially; then, the errors cease decreasing further, despite orders of magnitude increase in $N_f$ . This implies that unlike the conventional methods, where errors can be decreased as much as desired by following an instruction as simple as increasing the number of discretization points, in the theory-training, decreasing the errors is a far more challenging task. That challenge further highlights the importance of our guidelines that can help to decrease those errors, even though that decrease is
only one order of magnitude. Furthermore, recalling that the errors of more than seventy networks analyzed in Section \ref{section_res_b}  were all between $5\times 10^{-6}$ and $5\times 10^{-4}$ (in variables that are dimensionless and order unity) implies a disadvantage of the theory-training. Regardless of the number of training points, depth, or width of the network, theory-training errors cannot be made arbitrarily low. This disadvantage may limit the application of theory-training especially in domains where extremely low errors are required. Nonetheless, an error within the range we achieved here will be entirely acceptable for numerous engineering applications, such as macroscale modeling of solidification. In such applications, currently, the linear increase of the computational times with the number of
discretization points limits the simulation domain size and resolution, which results in simulations that have not fully resolved critical defects \cite{TMS}, forcing engineers to rely on simplified criterion \cite{Rayleigh} to predict them. If theory-training highly complex sets of equations that can predict those defects turns out to be successful, then the simulation domain size will not have to be limited by the available computational resources, and this is another area that requires further research.

\section{Conclusions}

By analyzing in detail the results of numerous computational experiments, we presented in-depth insights into theory-training neural networks to infer the solution of highly coupled differential equations. The results showed that the final training loss, which represents a TTN's ability to satisfy the equations at the training points, and the accuracy of the solution inferred at the new data points can be both deteriorated when the loss oscillates with training epochs. We presented a Partial
Regularization Training Technique (PRT) that is based on performing the training in two parts. In the first part, a regularized loss is minimized. After that loss has reached a plateau, training is continued by minimizing a non-regularized loss and ends after the latter loss has converged. The results showed that PRT eliminates the oscillations and, without increasing the training cost, gives a factor two improvement in the inference accuracy. That improvement was attributed to a factor ten decrease in the final training loss.

Among networks with different sizes, those that were deeper but not wider than the base TTN, which is the network with the lowest number of hidden layers and nodes per layer that is still able to decrease the training loss to reasonably low values, enjoyed a monotonous correlation between the training loss and inference error. The existence of such a correlation allowed us to present the following guidelines. In theory-training a set of equations, the base TTN should be identified first. Then, different TTNs that are all deeper but not wider than the base TTN need to be trained. Among the latter TTNs, the one with the lowest training loss, which is termed the optimal TTN, can be expected to have the lowest inference error, even when those errors cannot not be quantified. Following these guidelines will provide some confidence that the network ultimately used for inference has optimal training and inference times, as it does not have any additional unnecessary neurons, was trained to its full realizable learning capacity, and provides optimally accurate predictions.

A detailed comparison was made between theory-training and a rival, conventional method for solving differential equations that discretizes the derivatives. The results attested that even for a one-dimensional set of equations, theory-training can be as fast as conventional methods when the equations are highly coupled. This suggests that advantages of theory-training over the conventional methods are not necessarily limited to high-dimensional sets of equations and can be realized in modeling physical systems, where the dimensions are limited to four. Nonetheless, the comparison also revealed the limitation of theory-training: unlike conventional methods, in theory-training, errors cannot be made arbitrarily low, regardless of the number of training points, network depth, or width. This may limit the application of the present theory-training framework in domains where extreme accuracies are necessary.

It should be emphasized that our observations and guidelines were based only on the training experiments we performed here and lack a rigorous mathematical proof. Therefore, they are not guaranteed to hold if, for example, training is performed using an entirely different set of equations. Nonetheless, these observations can motivate further theoretical research on neural networks. For example, it would be interesting to see if the liner correlation between the training loss and inference error that we observed for networks that are deep but not wide has a theoretical reason and/or holds for other systems of equations.

\section{Acknowledgment}

This study was funded by the Deutsche Forschungsgemeinschaft (DFG, German Research Foundation) under Germany's Excellence Strategy - EXC-2023 Internet of Production-390621612. The authors appreciate fruitful discussions with Dr. Georg J. Schmitz of ACCESS e.V. MTR appreciates the comments by Dr. Ameya D. Jagtap of Brown University on regularization.

\bibliographystyle{plain}

\end{document}